\documentclass[preprint,authoryear,12pt]{elsarticle}

\usepackage{amssymb}
\usepackage{amsmath}
\usepackage{amsthm}
\usepackage{graphicx}
\usepackage{float}
\usepackage{placeins}
\usepackage{threeparttable}
\usepackage{verbatim}
\usepackage{bbding}
\usepackage{tabularray}
\usepackage[normalem]{ulem}
\usepackage[table,dvipsnames]{xcolor}
\usepackage{natbib}
\usepackage[colorlinks=true,citecolor=blue,linkcolor=blue,urlcolor=blue,breaklinks=true]{hyperref}

\usepackage{booktabs}
\usepackage{multirow}
\usepackage{makecell}
\usepackage{array}
\usepackage{algorithm}
\usepackage{algorithmic}
\usepackage{textcomp}
\usepackage{url}
\usepackage[final]{microtype}
\definecolor{avgrow}{RGB}{232,232,248}
\newcommand{\best}[1]{\textbf{#1}}
\newcommand{\second}[1]{\underline{#1}}

\journal{Medical Image Analysis}

\begin{document}

\begin{frontmatter}

\title{Towards Reliable Fetal Ultrasound Interpretation with Multi-Agent Collaboration}

\author[inst1]{Xiaotian Hu\fnref{fn1}}
\author[inst1]{Mingxuan Liu\fnref{fn1}}
\author[inst1]{Junwei Huang\fnref{fn1}}
\author[inst1]{Kasidit Anmahapong}
\author[inst1]{Yifei Chen}
\author[inst2]{Yiming Huang}
\author[inst1]{Xuguang Bai}
\author[inst1]{Zihan Li}
\author[inst1]{Hongjia Yang}
\author[inst1]{Yingqi Hao}
\author[inst3]{Hong Xu}
\author[inst3]{Yu Jiang}
\author[inst3]{Tian Tian}
\author[inst3]{Yi Liao}
\author[inst3]{Haibo Qu\corref{cor1}}
\author[inst1]{Qiyuan Tian\corref{cor1}}

\makeatletter
\emailauthor{qiyuantian@tsinghua.edu.cn}{Qiyuan Tian}
\emailauthor{windowsqhb@126.com}{Haibo Qu}
\makeatother

\affiliation[inst1]{organization={Tsinghua University},
            city={Beijing},
            country={China}}

\affiliation[inst2]{organization={University of California San Diego},
            city={La Jolla},
            state={CA},
            country={USA}}

\affiliation[inst3]{organization={West China Second University Hospital, Sichuan University},
            city={Chengdu},
            country={China}}

\fntext[fn1]{Xiaotian Hu, Mingxuan Liu, and Junwei Huang contributed equally to this work.}
\cortext[cor1]{Corresponding authors.}

\begin{abstract}
Automated fetal ultrasound interpretation spans from low-level visual perception (e.g., plane recognition and anatomical segmentation) to high-level clinical understanding (e.g., biometric measurement and diagnostic reporting). Deep learning models developed under the ``one-task, one-model'' paradigm impede systematic integration of perceptual evidence across the multi-step clinical workflow. Meanwhile, multimodal large language models (MLLMs) demonstrate promising visual understanding, yet limited domain-specific grounding and hallucination tendencies fundamentally constrain their reliability in fetal ultrasound analysis. To bridge this gap, we propose FetUSAgents, the first tool-augmented multi-agent system for comprehensive fetal ultrasound interpretation, supporting visual question answering (VQA), report generation, image captioning, and video summarization. FetUSAgents orchestrates task-specific visual tools via collaborative LLM agents, decomposing complex clinical queries into manageable subtasks that advance from anatomical recognition to quantitative measurement, mirroring expert sonographers' stepwise reasoning. Importantly, a Dual-Path Evidence Arbitration (DPEA) mechanism is proposed to integrate LLM-driven deliberative reasoning with structured computational evidence from specialized visual tools. Concurrently, a retrieval-enhanced Evidence bank consolidates heterogeneous intermediate findings to promote traceable, clinically grounded conclusions. Furthermore, we introduce FetUS-VQA, the first dedicated VQA benchmark for fetal ultrasound, comprising 1,892 images and 3,205 question–answer pairs spanning 10 clinical tasks. Extensive out-of-distribution (OoD) experiments demonstrate that FetUSAgents consistently surpasses both general and medical MLLMs, exceeding the strongest baseline by over 25\% in VQA accuracy, highlighting a scalable path toward evidence-driven clinical assistants for prenatal imaging. Code: \url{https://github.com/hu2274898/FetUSAgents}.
\end{abstract}


\end{frontmatter}

\section{Introduction}
Fetal ultrasound plays a central role in prenatal screening\citep{salomon2011routine} owing to its non-invasiveness, cost-effectiveness, and real-time imaging capability\citep{He2021PrenatalUltrasoundAI,Fiorentino2023DeepLearningFetalUltrasound}. In routine clinical practice, sonographers are required to identify standardized anatomical planes, delineate fine-grained fetal structures, and derive quantitative biometric measurements. However, these tasks must be performed within a dynamic scanning environment characterized by variable fetal positioning, acoustic shadowing, and limited soft-tissue contrast, rendering fetal ultrasound interpretation inherently subjective and highly operator-dependent\citep{Fetalclip}. In resource-limited settings, the scarcity of experienced sonographers further exacerbates these challenges, restricting timely access to reliable prenatal assessment \citep{Olagunju2025UltrasoundPregnancyOutcomes, sippel2011ultrasound}. Consequently, there is a pressing need for AI-assisted solutions to enhance the objectivity, reproducibility, and accessibility of fetal ultrasound diagnosis.

Prior deep-learning approaches for fetal ultrasound predominantly follow a ``one-task, one-model'' paradigm. Task-specific models for plane classification, anatomical structure segmentation, and fetal biometry have demonstrated promising accuracy on in-distribution (ID) benchmarks \citep{csm,AoP-SAM}. However, these models remain narrowly specialized and fragmented across workflows, typically deployed as standalone predictors that still require substantial manual coordination in clinical practice\citep{fiorentino2023review,yan2025aiultrasound}. Moreover, they are often developed on data from specific centers, specific devices, and narrowly defined task settings, which limits their robustness and transferability across institutions, imaging devices, populations, and low-resource clinical environments\citep{fiorentino2023review}.

Multimodal large language models (MLLMs)\citep{huatuo-gpt,medgemma}, pretrained or fine-tuned on large-scale medical datasets spanning diverse imaging modalities, have emerged as a promising approach to addressing these limitations. By learning joint visual-textual representations, MLLMs have demonstrated encouraging performance in medical visual question answering (VQA) and report generation. Nevertheless, they remain constrained by opaque reasoning processes that lack traceable visual evidence and cannot readily incorporate external domain knowledge\citep{EviAgent}. Moreover, most MLLMs rely on monolithic end-to-end inference, limiting their ability to invoke specialized tools or coordinate multi-step clinical workflows\citep{Meissa}. Their propensity for hallucination further poses substantial risks in safety-critical domains such as fetal ultrasound.

Tool-augmented agents alleviate the limited domain grounding, weak interpretability, and hallucination tendencies of end-to-end MLLMs by invoking external specialist tools and iteratively accumulating task-relevant evidence \citep{3DMedAgent}. Beyond tool use, LLM-driven agents introduce planning, memory, and role specialization, decomposing complex clinical queries into manageable subtasks while preserving intermediate evidence throughout the reasoning chain \citep{CT-Agent,WSI-Agent}. Collaborative multi-agent designs further enhance robustness through cross-verification and consultation-like deliberation, echoing multidisciplinary decision-making in clinical practice \citep{MD-Agent,MedAgents}. Nevertheless, existing agentic systems target primarily radiology, pathology, and general medical imaging, leaving fetal ultrasound unexplored. Given its distinctive challenges of dynamic scanning, subtle anatomical variations, and measurement-oriented interpretation, a dedicated agent-based framework tailored to this domain is still lacking.

To bridge this gap, we propose FetUSAgents, the first tool-augmented multi-agent system specifically designed for fetal ultrasound interpretation. Given a user query and ultrasound input, FetUSAgents routes the request to the appropriate clinical workflow, identifies the anatomical context, and dispatches task-specific expert Agents to acquire structured visual evidence. A Dual-Path Evidence Arbitration (DPEA) mechanism then integrates deliberative reasoning from multiple LLM-based voters with computational evidence from specialized tools, yielding decisions that are both clinically interpretable and data-grounded. A retrieval-enhanced Evidence bank further consolidates heterogeneous intermediate findings across the pipeline, strengthening factual grounding and reducing hallucination in the generated text. To systematically evaluate FetUSAgents, we constructed FetUS-VQA, the first comprehensive benchmark for fetal ultrasound, comprising 3,205 queries across 10 VQA subtasks derived from 7 clinical task categories (Fig. \ref{fig:crop_pic6}(a)). Extensive experiments on FetUS-VQA demonstrate that FetUSAgents consistently outperforms both general and medical MLLMs, exceeding the strongest competitor by over 25\% in VQA accuracy.

The main contributions of this work are as follows:
\begin{enumerate}
    \item We propose \textbf{FetUSAgents}, the first tool-augmented multi-agent system that enables general MLLMs to perform comprehensive fetal ultrasound interpretation, including VQA, report generation, and video summarization, without monolithic domain-specific retraining.
    
    \item We introduce a Dual-Path Evidence Arbitration (DPEA) mechanism and a retrieval-enhanced Evidence bank to improve decision reliability and effectively reduce hallucination in clinical report generation.

    \item We introduce FetUS-VQA, the first dedicated VQA benchmark for fetal ultrasound analysis. Extensive experiments on FetUS-VQA show that FetUSAgents consistently and significantly outperforms state-of-the-art (SOTA) general and medical MLLMs.
\end{enumerate}

\section{Related Work}

\subsection{Automatic Fetal Ultrasound Image Analysis}

Deep-learning methods have been widely explored for fetal ultrasound analysis, spanning standard-plane classification, anatomical structure segmentation, and fetal biometry. Specifically, early CNN-based models such as SonoNet enabled real-time detection and weakly supervised localization of fetal standard planes\citep{sononet}. Subsequent studies further advanced this task through systematic architecture evaluation\citep{fetalplanedb_paper}, deep feature integration\citep{cla_deepfeature}, and stacked ensemble learning\citep{cla_stack_ensemble}. Additionally, anatomical segmentation benefited from U-Net variants augmented with attention mechanisms for fetal cerebellum delineation\citep{ECAU-Net}, lightweight multi-scale designs for four-chamber heart segmentation\citep{mobilenet}, and semi-supervised strategies that alleviate pixel-level annotation demands\citep{Fetal-BCP}. Quantitative biometry has likewise progressed from segmentation-based geometric fitting for Head Circumference (HC) and Abdominal Circumference (AC), and femur length (FL) estimation\citep{AutoFB} toward landmark-based direct regression\citep{BiometryNet}, whole-examination frame-wise aggregation\citep{WholeExaminationAI}, and large-scale growth prediction beyond conventional Hadlock-style models\citep{AbnormalFetalGrowth}. More recently, foundation models such as FetalCLIP\citep{Fetalclip} and USFM\citep{USFM} have further advanced the field by enabling transferable representation learning across diverse downstream tasks. Despite these advances, most existing methods remain optimized for individual subtasks and lack the capacity to unify perceptual, quantitative, and interpretive stages into a coherent end-to-end clinical workflow.

\subsection{Medical Agentic Systems}
Medical agentic systems have emerged as a promising solution to the limitations of standalone MLLMs, demonstrating strong potential across clinical scenarios such as radiology\citep{CT-Agent,3DMedAgent,EviAgent}, pathology\citep{WSI-Agent}, and cardiology\citep{heartagent}. These systems typically coordinate multiple MLLM-based agents for debate, voting, or collaborative reasoning\citep{MedAgents,MD-Agent}, or integrate specialized medical tools to support structured evidence acquisition and multistep clinical decision-making\citep{medsamagent,Meissa}. However, to the best of our knowledge, existing agentic AI systems have not
yet been specifically adapted to fetal ultrasound, whose ambiguous anatomical
boundaries, limited soft-tissue contrast, operator-dependent image quality,
and scarcity of annotated data impose particularly unique requirements on
agent design.
\begin{figure}[!tbp]
\centering
\includegraphics[width=1.0\textwidth]{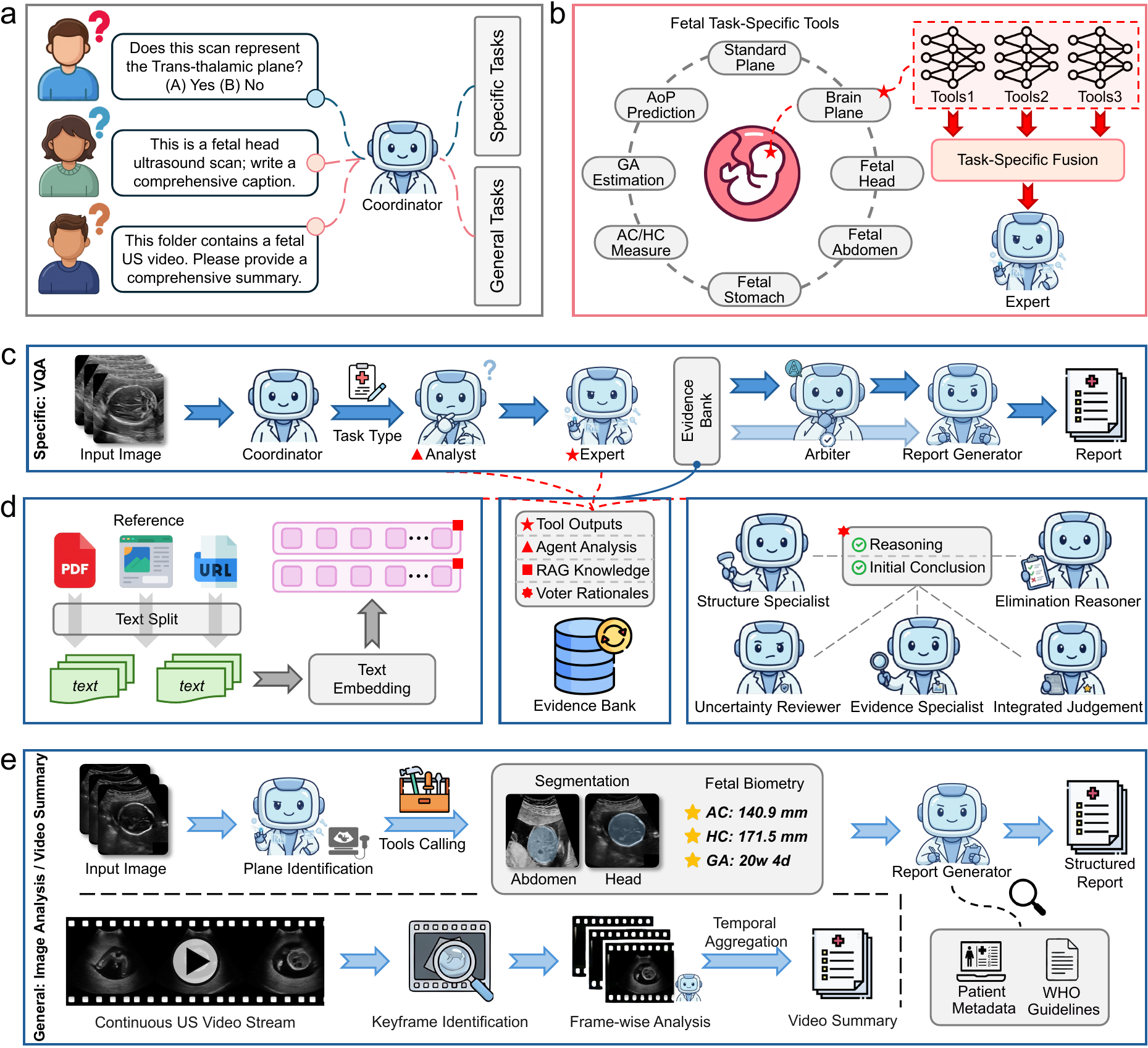}
\caption{Overview of the FetUSAgents framework. The system routes user requests to specific-task or general-task workflows, invokes task-specific Expert Agents for visual evidence acquisition, integrates tool outputs with multi-agent deliberation through DPEA, and uses retrieval-enhanced evidence consolidation to support grounded report generation, image captioning, and video summarization.}
\label{fig:crop_pic4}
\end{figure}

\section{Method}
\subsection{Overview}
FetUSAgents (Fig. \ref{fig:crop_pic4}) is a tool-augmented multi-agent system built upon the AutoGen framework\citep{autogen} for fetal ultrasound interpretation. Specifically, FetUSAgents comprises three core components: (1) a set of collaborative LLM-based agents that orchestrate task planning, deliberative reasoning, and report generation, unifying isolated visual predictors into a coherent clinical workflow (Sec.~B); (2) a task-specific tool library of specialized visual tools that serve as independently callable services, providing trustworthy anatomical
evidence for each clinical subtask (Sec.~C); and (3) a DPEA mechanism (Sec.~D) together with a retrieval-enhanced Evidence bank (Sec.~E) that integrates computational tool evidence with LLM-driven deliberative reasoning and consolidates heterogeneous intermediate findings to reduce hallucination in report generation. Orchestrating these components, given multimodal inputs including a fetal US image $I$, a video sequence $V = \{I_t\}_{t=1}^{T}$, and a user text query $Q$, FetUSAgents supports four clinical functions: VQA, report generation, image captioning, and video summarization.

\subsection{Agent Roles}

\subsubsection{Coordinator Agent}
The \textit{Coordinator Agent} (Fig.~\ref{fig:crop_pic4}(a)) serves as the central orchestrator of FetUSAgents, fulfilling two core responsibilities: (a) Query routing, which classifies user requests and dispatches them to the appropriate downstream workflow. Specifically, given a user query $Q$ and an input image $I$ or video $V$, the \textit{Coordinator Agent} determines the query type $\tau \in \{\textit{specific}, \textit{general}\}$. Queries with an explicit question-and-option structure (e.g., VQA) are categorized as \textit{specific}, whereas open-ended requests (e.g., image captioning and video summarization) are categorized as \textit{general}. Accordingly, the Coordinator dispatches \textit{specific} queries to the deliberative VQA pipeline (Sec.~D) and \textit{general} queries to the end-to-end clinical workflow (Sec.~F). (b) Task allocation, which identifies the clinical task type $t$ from inputs (i.e., $Q$, $I$), and activates the corresponding \textit{Expert Agent} (Sec.~B(3)) subset $E_{\mathrm{selected}} \subseteq \{E_1, \ldots, E_K\}$, which in turn invoke the relevant visual tools from the tool library (Sec.~C).

\subsubsection{Analyst Agent}
The \textit{Analyst Agent} parses the user query $Q$ and input image $I$ into a structured context representation $A$, which is shared with all downstream agents to support more focused reasoning.

\subsubsection{Expert Agents}
Each \textit{Expert Agent} $E_k$ is responsible for invoking a designated subset of visual tools
$S_k = \{s_{k,1}, \ldots, s_{k,N}\}$ from the task-specific tool library (Sec.~C) and applying a task-appropriate fusion rule $F_k$ (Sec.~C) to aggregate their outputs into a unified prediction:
\begin{equation}
\hat{o}_k = F_k\big(s_{k,1}(I), \ldots, s_{k,N}(I)\big),
\end{equation}
where $\hat{o}_k$ denotes the fused output of expert $E_k$. 

\subsubsection{Voter Agents}
To incorporate complementary, image-grounded clinical reasoning from diverse diagnostic perspectives, five deliberative \textit{Voter Agents} (Fig.~\ref{fig:crop_pic4}(d), right) were included, denoted as $\mathcal{V}=\{V_j\}_{j=1}^{5}$. Given an input image $I$ and a query $Q$, each voter $V_j$ independently evaluates the case under a distinct reasoning profile: (a)~\textit{Structure Specialist}, prioritizing anatomical identity recognition and key landmark localization; (b)~\textit{Evidence Specialist}, focusing on image appearance and morphological patterns; (c)~\textit{Elimination Reasoner}, adopting a differential-exclusion strategy that rules out implausible options first; (d)~\textit{Uncertainty Reviewer}, explicitly assessing evidence reliability and potential sources of ambiguity; and (e)~\textit{Integrated Judgement}, synthesizing all available cues into a globally coherent decision. This role diversification across the \textit{Voter Agents} encourages heterogeneous yet complementary evidence accumulation, thereby improving both the reliability and interpretability of the final decision.

\subsubsection{Arbiter Agent}
The \textit{Arbiter Agent} serves as the final decision-maker of FetUSAgents.
It integrates the evidence consolidated in the Evidence bank (Sec.~E)
and produces the final prediction~$\hat{y}$.

\subsubsection{Report Generator Agent}
The \textit{Report Generator Agent} synthesizes the evidence accumulated throughout the pipeline into a structured clinical report, as detailed in Sec.~E(3).

\subsection{Task-Specific Tool Library}
The task-specific tool library (Fig. \ref{fig:crop_pic4} B) covers three categories of fetal ultrasound analysis: plane classification, anatomical segmentation, and fetal biometry. \textit{Expert Agents} (Sec.~B) invoke the relevant tool subset and apply task-appropriate fusion rules to produce unified predictions.

\subsubsection{Plane Classification}
Accurate recognition of standard fetal planes is a prerequisite for reliable ultrasound interpretation. Four complementary classification tools for recognizing standard fetal views (i.e., head,
abdomen, femur, and thorax) as well as three fetal brain sub-planes (i.e., trans-cerebellar, trans-thalamic, and trans-ventricular) are provided:(i)~FetalCLIP\citep{Fetalclip}, a vision--language foundation model that provides robust representations for fetal ultrasound
images; (ii)~a ResNet-50 backbone initialized with RadImageNet\citep{RadImageNet} weights and equipped with a classification head; (iii)~FU-LoRA\citep{Fu-Lora}, a latent diffusion model fine-tuned via
Low-Rank Adaptation (LoRA)\citep{lora}; and
(iv)~a Vision Transformer (ViT-B/16)\citep{vit} initialized with ImageNet-1K\citep{imagenet} weights and equipped with a classification
head. The corresponding \textit{Expert Agent} applies agreement-based fusion for standard plane recognition and majority voting for brain sub-plane
classification.

\subsubsection{Anatomical Segmentation}
Fine-grained anatomical segmentation accurately localizes and delineates fetal structures, underpinning subsequent biometric measurement. The tool library provides segmentation tools organized by anatomical region. Specifically, for fetal head segmentation, three complementary tools are provided: (i) CSM\citep{csm}, a lightweight
CNN-based segmentation model; (ii) nnU-Net\citep{nnunet}, a self-configuring
medical image segmentation framework; and (iii) the Ultrasound Foundation Model (USFM)\citep{USFM}, adapted to this task by attaching and
fine-tuning a segmentation head. The \textit{Expert Agent} fuses the three mask outputs via pixel-level majority voting. For abdominal segmentation, two tools are provided: (i) FetalCLIP\citep{Fetalclip} and (ii) SAMUS\citep{SAMUS}, an interactive SAM-based ultrasound segmentation model. A sequential pipeline is employed as the fusion strategy: FetalCLIP first produces a coarse initial mask, from which the largest connected component and its centroid
are extracted to form the input prompt (bounding box and point) for SAMUS. For fetal stomach segmentation, three tools are provided: (i) nnU-Net\citep{nnunet}, (ii) FetalCLIP\citep{Fetalclip}, and (iii) SAMUS\citep{SAMUS}. The \textit{Expert Agent} applies mask-level majority voting, followed by ordered fallback when the fused mask is unreliable.

\subsubsection{Fetal Biometry}
Fetal biometry quantifies fetal growth and anatomy, underpinning developmental assessment and clinical decision-making. The tool library supports routine biometric tasks. Specifically, for angle of progression (AoP) measurement, three tools are provided: (i) AoP-SAM\citep{AoP-SAM}, a tailored SAM-based model for pubic symphysis and fetal head segmentation; (ii) USFM\citep{USFM}; and (iii) an adapted UperNet\citep{UperNet} model. The \textit{Expert Agent} applies a median-guided outlier-correction rule as the fusion strategy. For gestational age (GA) estimation, three tools are provided: (i)~FetalCLIP\citep{Fetalclip}, along with two
RadImageNet\citep{RadImageNet}-initialized backbones equipped with classification heads, namely (ii)~ResNet-50 and
(iii)~ConvNeXt-Tiny\citep{convnext}. The \textit{Expert Agent} integrates the tool predictions using a consistency-aware weighted strategy. For HC and AC measurements, no additional dedicated tools are required. HC and AC are deterministically derived by ellipse fitting from the high-fidelity segmentation masks produced by the head and abdominal segmentation tools. The \textit{Expert Agent} applies a priority-based fallback strategy to ensure robust HC and AC measurements when a preferred tool fails.

\subsection{Deliberative VQA for Specific Tasks}
The \textit{Specific Tasks} branch handles fetal ultrasound VQA queries 
with explicit questions and candidate options 
(Fig.~\ref{fig:crop_pic4}(c)), e.g., \textit{Does this scan represent the trans-thalamic plane? 
(A) Yes (B) No}. The \textit{Coordinator Agent} (Sec.~B) routes such queries to this workflow, identifies the task type $t$, and 
selects the corresponding expert subset 
$E_{\mathrm{selected}}$ for tool invocation. The \textit{Analyst Agent} (Sec.~B) then transforms the user query $\mathcal{Q}$, 
candidate option set $\mathcal{O}$, and input image $I$ into 
a structured question context shared with downstream agents:
\begin{equation}
A = \mathrm{Analyst}(Q, \mathcal{O}, I).
\end{equation}

To combine interpretable reasoning evidence with reliable tool-supported evidence, a DPEA mechanism is proposed, comprising a deliberative path that captures diverse diagnostic reasoning from multiple LLM-based voters, and a computational path that provides structured predictions from task-specific tools. 

In the deliberative path, five \textit{Voter Agents} with distinct reasoning profiles (Sec.~B) are deployed. This design is inspired by multidisciplinary team (MDT) consultation in clinical practice, where complementary expert perspectives are integrated for case-level decision making \citep{mdt}. Each voter independently examines the same case and produces both an option prediction and a textual rationale, thereby yielding diverse yet complementary reasoning evidence. Specifically, each voter independently produces a prediction--rationale pair, and the complete deliberative evidence set is:
\begin{equation}
E_{\mathrm{vote}} = \{(l_j, r_j)\}_{j=1}^{5},\quad 
(l_j, r_j) = V_j(Q, \mathcal{O}, I, A),
\end{equation}
where $l_j \in \mathcal{O}$ denotes the predicted option label and 
$r_j$ the corresponding rationale.

In parallel, the computational path invokes the selected task-specific 
\textit{Expert Agents} according to task type $t$ and performs precise 
tool-based inference on the input image. Each expert communicates with 
the rest of the system exclusively through structured JSON outputs, 
thereby reducing subjective free-form intermediate generation and 
improving factual reliability. Specifically, the complete computational evidence set is:
\begin{equation}
E_{\mathrm{tool}} = \{(\hat{o}_k, \mathcal{D}_k)\}_{E_k \in 
E_{\mathrm{selected}}},\quad 
(\hat{o}_k, \mathcal{D}_k) = E_k(I),
\end{equation}
where $\hat{o}_k$ denotes the fused prediction of expert $E_k$ and 
$\mathcal{D}_k$ the associated decision details (e.g., measurements 
or intermediate structured evidence).

The final prediction is produced by the \textit{Arbiter Agent}, which 
also receives the aggregated Evidence bank $M$ (Sec.~E). By jointly evaluating the consistency, 
complementarity, and relative strength of the two evidence paths, the 
\textit{Arbiter Agent} ensures that neither pure language reasoning nor pure 
computational prediction dominates in isolation, yielding decisions 
that are both data-grounded and clinically interpretable.

\subsection{Evidence Integration and Report Generation}
Reliable fetal ultrasound report generation requires coherent 
integration of multi-source clinical evidence, multi-agent reasoning, 
and external medical knowledge. To this end, FetUSAgents combines 
retrieval-augmented external knowledge with a unified Evidence bank 
that consolidates heterogeneous evidence across the pipeline 
(Fig.~\ref{fig:crop_pic4}(d)), serving both the specific-task 
(Sec.~D) and general-task (Sec.~F) workflows.

\subsubsection{Retrieval-Augmented Generation}
Retrieval-augmented generation (RAG)\citep{rag} (Fig.~\ref{fig:crop_pic4}(d), left) is incorporated to provide external medical evidence. Specifically, an external knowledge base is curated from authoritative references, 
including WHO guidelines\citep{WHO} and ISUOG practice standards, which 
provide guidance for standard plane acquisition, anatomical 
landmark identification, biometric measurement, and clinical interpretation in fetal ultrasound. All reference documents are split into semantically coherent chunks 
$K = \{k_i\}_{i=1}^{N}$ and embedded into a shared vector space via a 
pretrained encoder $f_{\theta}(\cdot)$. Given a query $Q$, the system 
retrieves the top-$k$ most relevant snippets as external medical 
evidence:
\begin{equation}
E_{\mathrm{rag}} = \mathrm{Retrieve}\bigl(Q,\; K\bigr) 
= \{k_i\}_{i=1}^{k},
\end{equation}
which informs both the \textit{Arbiter Agent}'s decision and the subsequent report 
generation.

\subsubsection{Evidence bank}
In a multi-agent pipeline, clinically meaningful reports must coherently
integrate anatomical observations, quantitative measurements,
multi-agent reasoning, and external medical knowledge. However, these
evidence sources are distributed across heterogeneous modules and may
be only partially preserved during free-form text synthesis. To address
this issue, FetUSAgents introduces a unified Evidence bank to aggregate
intermediate evidence from multiple sources:
\begin{equation}
M = \{A, E_{\mathrm{vote}}, E_{\mathrm{tool}}, E_{\mathrm{rag}}\}.
\end{equation}
For the specific-task workflow, all four components are populated as
described in Sec.~D and Sec.~E (1). For the general-task workflow
(Sec.~F), where no explicit candidate options or voter deliberation are
involved, the Evidence bank reduces to
\begin{equation}
M_{\mathrm{gen}} = \{A, E_{\mathrm{tool}}, E_{\mathrm{rag}}\},
\end{equation}
retaining the tool evidence and retrieved knowledge to support
report generation.

\subsubsection{Report Generation}
Based on the aggregated evidence, the \textit{Report Generator Agent} 
consolidates the Evidence bank $M$ together with the final prediction 
$\hat{y}$ from the \textit{Arbiter Agent} to produce the structured 
clinical report, which comprises three sections: (i) \textit{Findings}, which 
objectively describe observable anatomical structures and image 
features; (ii) \textit{Impression}, which summarizes the final clinical 
interpretation; and (iii) \textit{Note}, which provides a summary 
of the supporting evidence.

\subsection{End-to-End Clinical Workflow for General Tasks}

The \textit{General Tasks} branch addresses open-ended requests
(Fig.~\ref{fig:crop_pic4}(e)) and covers two sub-tasks:
\textit{image caption generation}
(e.g., \textit{``Generate a caption for this ultrasound image''})
and \textit{video summarization}
(e.g., \textit{``Summarize the key findings in this ultrasound video''}).
Given a user query $Q$, once the \textit{Coordinator Agent} routes
the case to this branch, it further identifies which sub-task
the request belongs to.

\subsubsection{Image Caption Generation.}
For open-ended image queries, FetUSAgents identifies the anatomical
plane and activates the corresponding \textit{Expert Agents} to
extract anatomical observations and quantitative measurements. Beyond this standard pipeline, a WHO-chart-based internal consistency check \citep{WHO} is introduced to prevent contradictory reporting by correlating derived biometric indices (e.g., HC, AC) with the estimated GA to compute normative growth percentiles. When a measurement falls outside clinically plausible ranges, a reflection step re-evaluates the expert outputs and replaces
unreliable estimates with robust ensemble fallbacks. The verified findings are then consolidated into a structured report via \textit{Report Generator Agent} with the Evidence bank.

\subsubsection{Video Summarization.}
In a continuous ultrasound stream $V{=}\{I_t\}_{t=1}^{T}$, only a
small fraction of frames contain standardized diagnostic views.
FetUSAgents deploys a keyframe extractor based on
FetalCLIP \citep{fetalclipiqa} to automatically isolate and classify
clinically meaningful planes, which are then dispatched to the
corresponding \textit{Expert Agents} for frame-wise analysis.
The resulting multi-frame evidence is aggregated via the Evidence
Bank and \textit{Report Generator Agent} into a cohesive, sequence-level
clinical summary that captures key anatomical and biometric
findings across the entire examination.
This pipeline mirrors the real-world sonographer workflow of
scanning, identifying key views, measuring, and synthesizing
findings, thereby offering a clinically practical solution for
open-ended fetal ultrasound interpretation.

\section{Experiment}

\subsection{Datasets and FetUS-VQA Construction}

\subsubsection{Overview}
\begin{figure}[!tbp]
\centering
\includegraphics[width=1.0\textwidth]{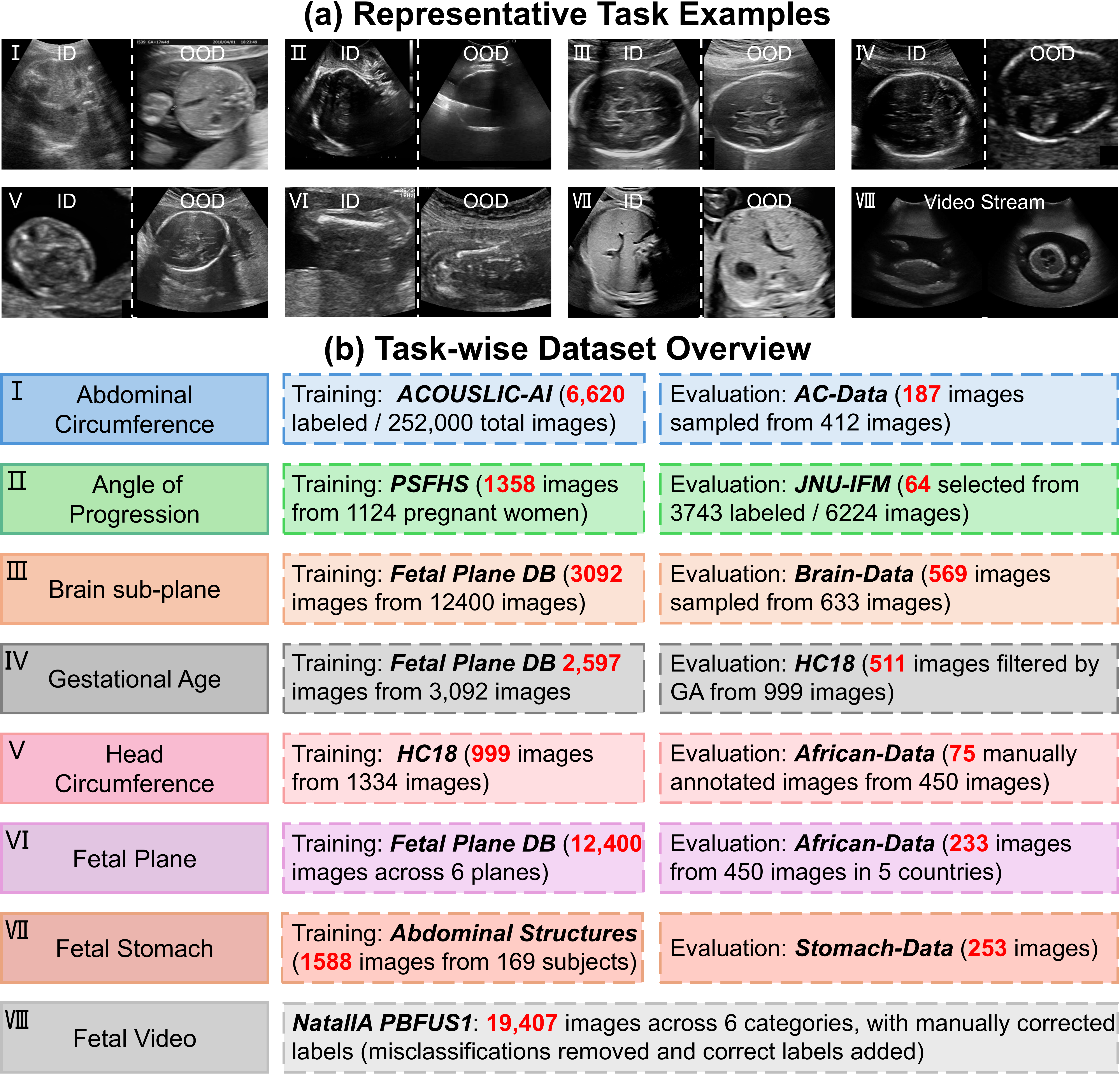}
\caption{Overview of datasets and tasks. (a) Representative in-distribution (ID) and out-of-distribution (OoD) image pairs for each task. (b) Task-wise summary of training and evaluation datasets.}
\label{fig:crop_pic6}
\end{figure}

Our experiments employ 10 publicly available datasets covering 7 clinically relevant tasks
(Fig.~\ref{fig:crop_pic6}). For each task, one dataset is used to
train the visual tools within the Expert Agents, while an
independently collected counterpart is reserved for OoD evaluation,
ensuring cross-institutional generalization. The OoD evaluation sets
additionally serve as the image source for the FetUS-VQA (Sec.~\ref{sec:vqa_construction}).

\begin{figure}[!tbp]
\centering
\includegraphics[width=1.0\textwidth]{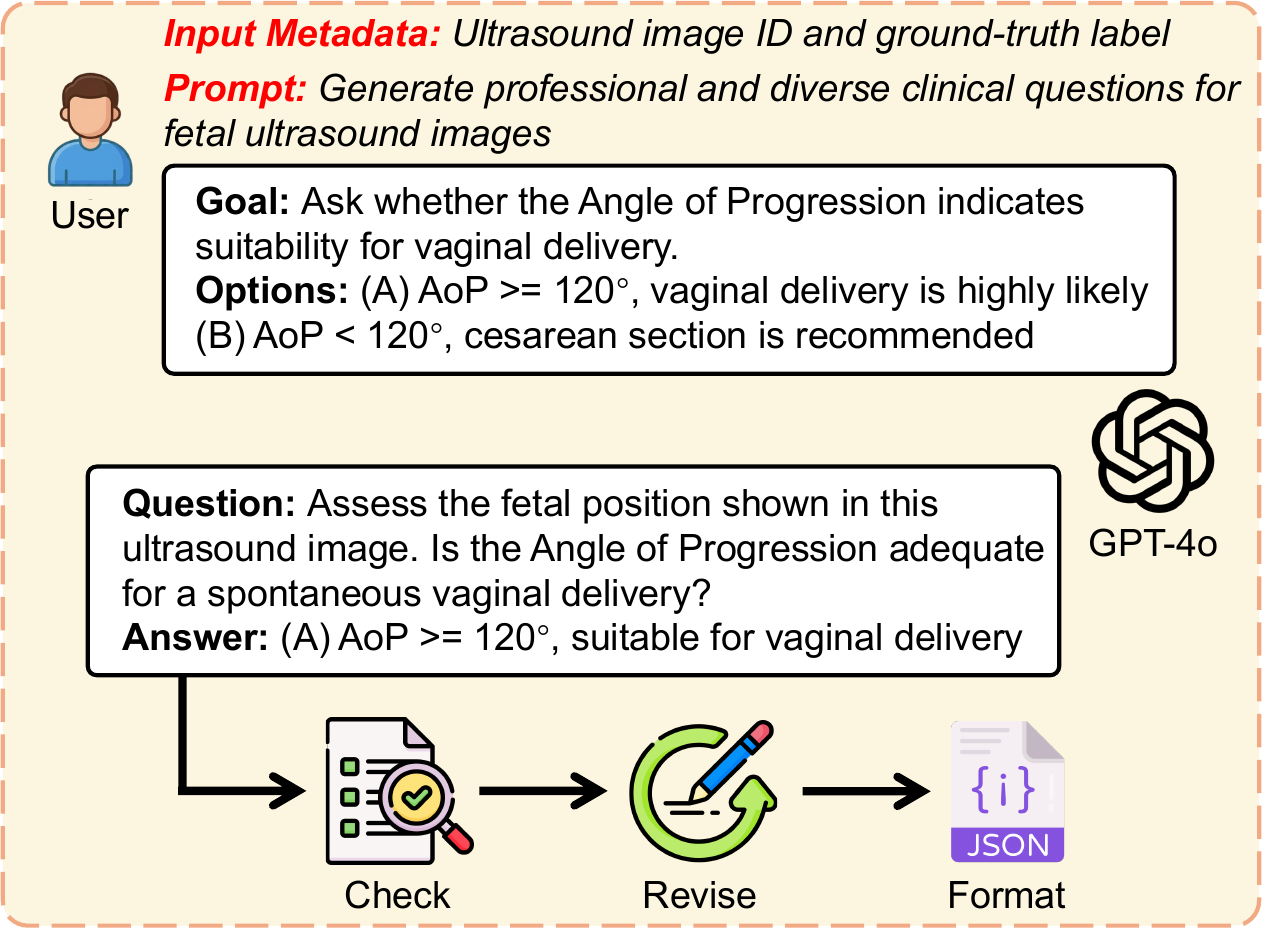}
\caption{Semi-automated pipeline for FetUS-VQA construction. Given an image ID and its ground-truth label, a task-aware prompt guides GPT-4o to generate clinically grounded question--answer pairs, which then undergo expert checking, revision, and standardized JSON formatting.}
\label{fig:crop_pic1}
\end{figure}

\subsubsection{Task-Specific Datasets}

(i) \textit{Abdominal Circumference (AC) Estimation}: The ACOUSLIC-AI \citep{ACOUSLIC-AI} dataset was used for training, which contains 252{,}000 blind-sweep ultrasound images, among which 6{,}620 images with abdominal annotations were selected. For OoD evaluation, AC-Data \citep{acdata} provides 412 images, from which 187 images were selected. (ii) \textit{Angle of Progression (AoP Estimation)}: The PSFHS \citep{psfhs} dataset (1{,}358 images) was used for training. For OOD evaluation, 64 labeled images from different videos in JNU-IFM \citep{JNU} (6{,}224 images, 3{,}743 labeled) were used. (iii) \textit{Brain Sub-plane Classification}: The brain subset of Fetal Plane DB \citep{fetalplanedb} (3{,}092 images: 1{,}638 trans-thalamic, 714 trans-cerebellar, 597 trans-ventricular, and 143 other) was used for training. For OOD evaluation, 569 selected images (90 trans-cerebellar, 228 trans-thalamic, 251 trans-ventricular) from 633 Brain-Data \citep{brain+stomach} images were used. (iv) \textit{Gestational Age (GA) Estimation}: The Fetal Plane DB \citep{fetalplanedb} dataset (2{,}597 brain images) was used for training, and 511 brain images from HC18 \citep{hc} were used for OOD evaluation. As neither dataset provides GA annotations, following FetalCLIP \citep{Fetalclip}, HC is first computed from the segmentation masks and then converted to GA via numerical inversion of the WHO fetal growth model. (v) \textit{Head Circumference (HC) Estimation}: The HC18 \citep{hc} dataset (1{,}334 images total) was used for training, from which 999 brain images were selected. For OOD evaluation, 75 images with manual annotations\citep{wangfangdata} were selected from the African-Data \citep{African} dataset. (vi) \textit{Plane Classification}:
The full Fetal Plane DB \citep{fetalplanedb} dataset
(12{,}400 images: 711 abdomen, 3{,}092 brain, 1{,}040 femur,
1{,}718 thorax, 1{,}626 cervix, and 4{,}213 other) was used
for training.
For OoD evaluation, the African-Data \citep{African} dataset (125 abdomen, 125 brain, 125 femur, and 75 thorax) was used. (vii) \textit{Stomach Segmentation}: The Abdominal Structures \citep{brain+stomach} dataset (1{,}588 images) was used for training and Stomach-Data \citep{stomach} (253 images) was used for OOD evaluation.

\subsubsection{Video Summarization Dataset}
For video-level summarization evaluation, the public fetal ultrasound video dataset Natalia PBFUS1 \citep{video} is adopted, containing 19{,}407 frames in total. The original frame-level annotations were revised by one experienced physician, yielding 47 Biparietal Plane, 78 Abdominal Plane, 68 Heart Plane, 139 Spine Plane, 47 Femur Plane, and 19{,}028 No Plane frames.

\begin{table}[t]
\centering
\caption{Overview of the FetUS-VQA benchmark.}
\label{tab:dataset_overview}
\setlength{\tabcolsep}{4pt}
\renewcommand{\arraystretch}{1.18}
\begin{tabular*}{\columnwidth}{@{\extracolsep{\fill}}llccc}
\toprule
\multicolumn{5}{c}{\cellcolor{gray!15}\textbf{Dataset Overview}} \\
\midrule
\multicolumn{4}{l}{Fetal ultrasound images} & 1{,}892 \\
\multicolumn{4}{l}{VQA pairs} & 3{,}205 \\
\midrule
\multicolumn{5}{c}{\cellcolor{gray!15}\textbf{Question Taxonomy}} \\
\midrule
\textbf{Category} & \textbf{ID} & \textbf{Task} & \textbf{Options} & \textbf{VQA pairs} \\
\midrule

\multirow{4}{*}{Measurement}
& Task1  & AC           & 4 (A/B/C/D) & 187 \\
& Task2  & AoP          & 2 (A/B)     & 64  \\
& Task9  & HC           & 4 (A/B/C/D) & 75  \\
& Task10 & Stomach      & 4 (A/B/C/D) & 253 \\
\cmidrule(lr){2-5}
& \textbf{Subtotal} & \textbf{4 tasks} &  & \textbf{579} \\
\midrule

\multirow{6}{*}{Classification}
& Task3  & Brain Binary & 2 (A/B)     & 569 \\
& Task4  & Brain Multi  & 3 (A/B/C)   & 569 \\
& Task5  & Plane Binary & 2 (A/B)     & 233 \\
& Task6  & Plane Multi  & 4 (A/B/C/D) & 233 \\
& Task7  & GA Binary    & 2 (A/B)     & 511 \\
& Task8  & GA Multi     & 3 (A/B/C)   & 511 \\
\cmidrule(lr){2-5}
& \textbf{Subtotal} & \textbf{6 tasks} &  & \textbf{2{,}626} \\
\midrule

\textbf{Total} &  & \textbf{10 tasks} &  & \textbf{3{,}205} \\
\bottomrule
\end{tabular*}
\end{table}

\subsubsection{FetUS-VQA Construction}
\label{sec:vqa_construction}
To enable end-to-end assessment of multimodal reasoning, FetUS-VQA was constructed from the OoD evaluation datasets as the first VQA benchmark dedicated to fetal ultrasound. It comprises 1{,}892 images and 3{,}205 question–answer pairs across 10 clinical tasks (Table~\ref{tab:dataset_overview}). Specifically, FetUS-VQA is built via a semi-automated pipeline (Fig.~\ref{fig:crop_pic1}). For each image, its identifier and ground-truth label are fed into a task-aware prompt that guides GPT-4o to generate clinically grounded question--answer pairs. For classification tasks, the label serves as the correct option with the remaining class labels as distractors. Each such task is further formulated in both binary (e.g., \textit{``Does this scan represent the trans-thalamic plane? (A)~Yes (B)~No''}) and multi-class (e.g., \textit{``Identify the cranial plane in this fetal brain ultrasound: (A)~Trans-cerebellum, (B)~Trans-thalamic, (C)~Trans-ventricular''}) formats to probe robustness under varying diagnostic difficulty. For measurement tasks, a quantitative value (e.g., circumference in mm, angle in degrees, or area in cm\textsuperscript{2}) is computed from the segmentation mask and discretized into candidates at clinically meaningful intervals. All generated pairs undergo expert review to ensure clinical accuracy and answer validity, and the finalized benchmark is released in standardized JSON format.

\subsection{Comparative MLLMs}
For comprehensive evaluation, we benchmark FetUSAgents against a diverse set of baseline MLLMs, including both general models (i.e., GPT-5.1\citep{gpt5.1}, GPT-5.4\citep{gpt5.4}, Qwen3-Max\citep{qwen3max}, Qwen3-VL-30B-A3B\citep{qwen3vl}, DeepSeek-V3.2\citep{deepseekv3.2}, Doubao-Seed-2.0-Pro\citep{doubao2.0pro}, and Gemini-3-Flash-Preview-Thinking\citep{gemini3flash}) and medical models (i.e., LLaVA-Med-v1.5-Mistral-7B\citep{llava-med}, HuatuoGPT-34B\citep{huatuo-gpt}, Hulu-Med-30A3\citep{hulumed}, Hulu-Med-32B\citep{hulumed}, and MedGemma-27B\citep{medgemma}).

\subsection{Implementation Details}
All experiments were conducted on a workstation equipped with a single NVIDIA A800 GPU with 80 GB VRAM. All LLM-driven agents in FetUSAgents used GPT-5.1\citep{gpt5.1} as the language-model backbone. For RAG, a knowledge base was constructed using Chroma as the persistent vector database. Reference documents were embedded using OpenAIEmbeddings\citep{openaiembeddings} with the text-embedding-3-small\citep{text_embedding_3_small} embedding model. During inference, FetUSAgents retrieved the top 5 most relevant knowledge snippets for each query.

\begin{figure}[!t]
\centerline{\includegraphics[width=\columnwidth]{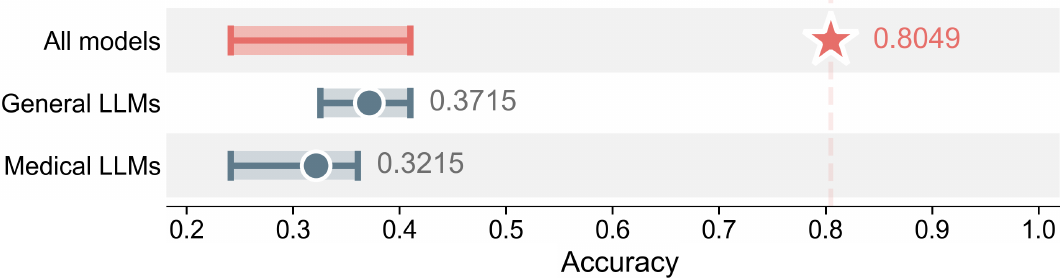}}
\caption{VQA accuracy on measurement tasks. Circles: group means; bars: ranges; stars: FetUSAgents.}
\label{fig:crop_pic8}
\end{figure}
\begin{figure}[!t]
\centerline{\includegraphics[width=\columnwidth]{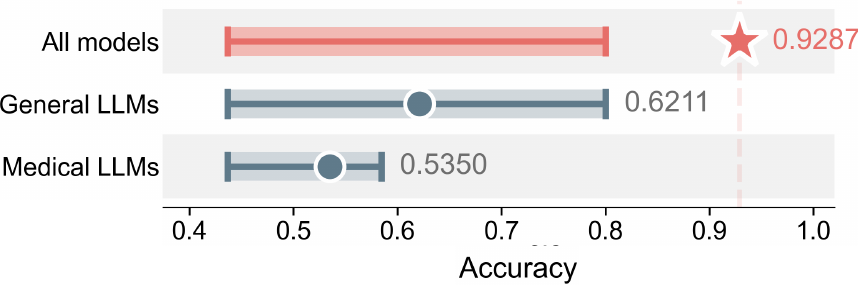}}
\caption{VQA accuracy on classification tasks. Circles: group means; bars: ranges; stars: FetUSAgents.}
\label{fig:crop_pic7}
\end{figure}

\subsection{Comparison with MLLMs on FetUS-VQA}

FetUSAgents consistently outperformed all general and medical MLLMs on the FetUS-VQA benchmark in both accuracy and macro-F1 score (Tables~\ref{tab:vqa_accuracy} and \ref{tab:vqa_f1}). Specifically, FetUSAgents achieved an average accuracy of 0.8791 and an average macro-F1 score of 0.8721, surpassing the strongest general baseline, Gemini-3-Flash-Thinking (accuracy, 0.6272; macro-F1, 0.5509), by 25.19\% and 32.12\%, and the best medical baseline, Hulu-Med-32B (accuracy, 0.4941; macro-F1, 0.4493), by 38.50\% and 42.28\%, with this superiority maintained across all ten tasks. FetUSAgents also exhibited well-balanced class-wise predictions, with closely aligned accuracy and macro-F1 scores, whereas most baselines showed lower macro-F1 than accuracy, indicating biased category preferences. Notably, when accuracy is disaggregated by task category (Figs.~\ref{fig:crop_pic8} and \ref{fig:crop_pic7}), baseline MLLMs scored substantially lower on measurement tasks (0.2--0.5) than on classification tasks (0.4--0.8), revealing that current models remain more capable of semantic recognition than geometric quantification. FetUSAgents, by contrast, exceeded 0.80 accuracy on both measurement (0.8049) and classification (0.9287) tasks, effectively bridging this gap. General MLLMs consistently outperformed their medical counterparts, suggesting that fine-tuning with limited medical corpora does not necessarily confer advantages in fetal ultrasound.

\begin{table}[t]
\centering
\caption{VQA accuracy (\%) on the FetUS-VQA benchmark. Best results are in \textbf{bold} and runner-up \underline{underlined}. ``Rand'': the random-guess baseline.}
\label{tab:vqa_accuracy}
\setlength{\tabcolsep}{4pt}
\renewcommand{\arraystretch}{1.12}
\resizebox{\textwidth}{!}{%
\begin{tabular}{@{}lccccccccccc@{}}
\toprule

\textbf{Method} & \textbf{Task1} & \textbf{Task2} & \textbf{Task3} & \textbf{Task4} & \textbf{Task5} & \textbf{Task6} & \textbf{Task7} & \textbf{Task8} & \textbf{Task9} & \textbf{Task10} & \textbf{Avg.} \\
\midrule
\textcolor{gray}{\textbf{Rand}} & \textcolor{gray}{0.2500} & \textcolor{gray}{0.5000} & \textcolor{gray}{0.5000} & \textcolor{gray}{0.3333} & \textcolor{gray}{0.5000} & \textcolor{gray}{0.2500} & \textcolor{gray}{0.5000} & \textcolor{gray}{0.3333} & \textcolor{gray}{0.2500} & \textcolor{gray}{0.2500} & \textcolor{gray}{0.3667} \\ 
\midrule
\rowcolor{gray!15}
\multicolumn{12}{c}{\textbf{(a) General large language models}} \\
GPT-5.1 & 0.3048 & 0.5469 & 0.5501 & 0.3533 & 0.7425 & 0.6781 & 0.8160 & 0.6986 & 0.3867 & 0.2767 & 0.5354 \\
GPT-5.4 & 0.3262 & 0.4688 & 0.5641 & 0.4112 & 0.7854 & 0.7511 & 0.8043 & \underline{0.8278} & 0.4000 & 0.1067 & 0.5446 \\
Qwen3-MAX         & 0.4439 & 0.5938 & 0.5009 & 0.3831 & 0.4936 & 0.2790 & 0.2505 & 0.7123 & 0.2933 & 0.3083 & 0.4259 \\
Qwen3-VL-30B-A3B        & \second{0.4599} & 0.4062 & 0.5026 & 0.4183 & 0.6266 & 0.4893 & 0.4090 & 0.7456 & 0.3333 & 0.3320 & 0.4723 \\
DeepSeek-V3.2                     & 0.2995 & 0.4531 & 0.5009 & 0.4007 & 0.4807 & 0.2790 & 0.7456 & 0.7358 & 0.2933 & 0.2767 & 0.4465 \\
Doubao-Seed-2.0-Pro                   & 0.4492 & 0.5312 & 0.6204 & 0.5817 & 0.8240 & 0.7639 & \underline{0.8356} & 0.7241 & 0.3333 & 0.3083 & 0.5972 \\
Gemini-3-Flash-Thinking      & 0.3209 & 0.5000 & \second{0.7100} & \underline{0.7311} & \underline{0.9185} & \underline{0.9056} & 0.7828 & 0.7534 & \underline{0.4400} & 0.2095 & 0.6272 \\
\midrule

\rowcolor{gray!15}
\multicolumn{12}{c}{\textbf{(b) Medical large language models}} \\
Llava-Med-V1.5-Mistral-7B       & 0.1979 & 0.5938 & 0.4148 & 0.4411 & 0.4893 & 0.2747 & 0.2544 & 0.7456 & 0.2400 & 0.2688 & 0.3960 \\
HuatuoGPT-34B   & 0.1925 & 0.3594 & 0.5413 & 0.4341 & 0.6438 & 0.3734 & 0.3659 & 0.6301 & 0.2000 & 0.2134 & 0.3954 \\
Hulu-Med-30A3         & 0.3690 & 0.4375 & 0.5888 & 0.4060 & 0.6309 & 0.7167 & 0.3875 & 0.7456 & 0.2400 & 0.2411 & 0.4763 \\
Hulu-Med-32B         & 0.3422 & 0.5626 & 0.5975 & 0.4183 & 0.7039 & 0.6009 & 0.4364 & 0.7495 & 0.2133 & \underline{0.3162} & 0.4941 \\
MedGemma-27B                    & 0.2567 & \second{0.6250} & 0.5729 & 0.4464 & 0.5708 & 0.3777 & 0.7456 & 0.7456 & 0.2800 & 0.2806 & 0.4901 \\
\midrule   

\textbf{FetUSAgents} & \best{0.9947} & \best{0.7188} & \best{0.9332} & \best{0.9086} & \best{0.9399} & \best{0.9313} & \best{0.9295} & \best{0.9295} & \best{1.0000} & \best{0.5059} & \best{0.8791} \\
\bottomrule
\end{tabular}%
}
\end{table}

\begin{table}[t]
\centering
\caption{VQA macro-F1 score on the FetUS-VQA benchmark. Best results are in \textbf{bold} and runner-up \underline{underlined}.}
\label{tab:vqa_f1}
\setlength{\tabcolsep}{4pt}
\renewcommand{\arraystretch}{1.12}
\resizebox{\textwidth}{!}{%
\begin{tabular}{@{}lccccccccccc@{}}
\toprule

\textbf{Method} & \textbf{Task1} & \textbf{Task2} & \textbf{Task3} & \textbf{Task4} & \textbf{Task5} & \textbf{Task6} & \textbf{Task7} & \textbf{Task8} & \textbf{Task9} & \textbf{Task10} & \textbf{Avg.} \\
\midrule
\rowcolor{gray!15}
\multicolumn{12}{c}{\textbf{(a) General large language models}} \\
GPT-5.1 & 0.2880 & 0.5205 & 0.5484 & 0.3401 & 0.7419 & 0.6420 & 0.7498 & 0.3207 & 0.3434 & 0.2560 & 0.4751 \\
GPT-5.4 & 0.3165 & 0.4688 & 0.5500 & 0.3866 & 0.7805 & 0.7176 & \underline{0.7685} & \underline{0.7033} & 0.3261 & 0.1021 & 0.5120 \\
Qwen3-MAX         & \underline{0.4297} & 0.3725 & 0.3368 & 0.2416 & 0.4070 & 0.1094 & 0.2020 & 0.2789 & 0.2651 & 0.2590 & 0.2902 \\
Qwen3-VL-30B-A3B        & 0.4073 & 0.3091 & 0.3928 & 0.2980 & 0.6194 & 0.3590 & 0.4026 & 0.4271 & 0.2975 & 0.2789 & 0.3792 \\
DeepSeek-V3.2                     & 0.2449 & 0.4465 & 0.3337 & 0.1976 & 0.4782 & 0.1630 & 0.4271 & 0.2826 & 0.1947 & 0.2148 & 0.2983 \\
Doubao-Seed-2.0-Pro                   & 0.4078 & 0.5271 & 0.6202 & 0.4638 & 0.8197 & 0.6906 & 0.7285 & 0.2901 & 0.3321 & \underline{0.2968} & 0.5177 \\
Gemini-3-Flash-Thinking      & 0.3148 & 0.3592 & \textit{0.7012} & \underline{0.6893} & \underline{0.9184} & \underline{0.8803} & 0.5779 & 0.4589 & \underline{0.4149} & 0.1939 & \underline{0.5509} \\
\midrule

\rowcolor{gray!15}
\multicolumn{12}{c}{\textbf{(b) Medical large language models}} \\
Llava-Med-V1.5-Mistral-7B       & 0.1796 & 0.3725 & 0.4071 & 0.2041 & 0.3378 & 0.1747 & 0.2028 & 0.4271 & 0.1253 & 0.1182 & 0.2549 \\
HuatuoGPT-34B   & 0.1789 & 0.3580 & 0.4781 & 0.2918 & 0.6431 & 0.2753 & 0.3591 & 0.5679 & 0.1771 & 0.2043 & 0.3534 \\
Hulu-Med-30A3         & 0.2944 & 0.4325 & 0.5887 & 0.1995 & 0.5830 & 0.6345 & 0.3874 & 0.4271 & 0.1556 & 0.1557 & 0.3858 \\
Hulu-Med-32B         & 0.3248 & \underline{0.5586} & 0.5601 & 0.2762 & 0.7038 & 0.5407 & 0.4364 & 0.5928 & 0.2042 & 0.2949 & 0.4493 \\
MedGemma-27B                    & 0.1762 & 0.5194 & 0.5691 & 0.2632 & 0.5690 & 0.3054 & 0.4271 & 0.4271 & 0.1491 & 0.2284 & 0.3634 \\
\midrule

\textbf{FetUSAgents} & \textbf{0.9942} & \textbf{0.7046} & \textbf{0.9332} & \textbf{0.8958} & \textbf{0.9396} & \textbf{0.9277} & \textbf{0.9102} & \textbf{0.9102} & \textbf{1.0000} & \textbf{0.5051} & \textbf{0.8721} \\
\bottomrule
\end{tabular}%
}
\end{table}

\begin{figure}[!t]
\centerline{\includegraphics[width=\textwidth]{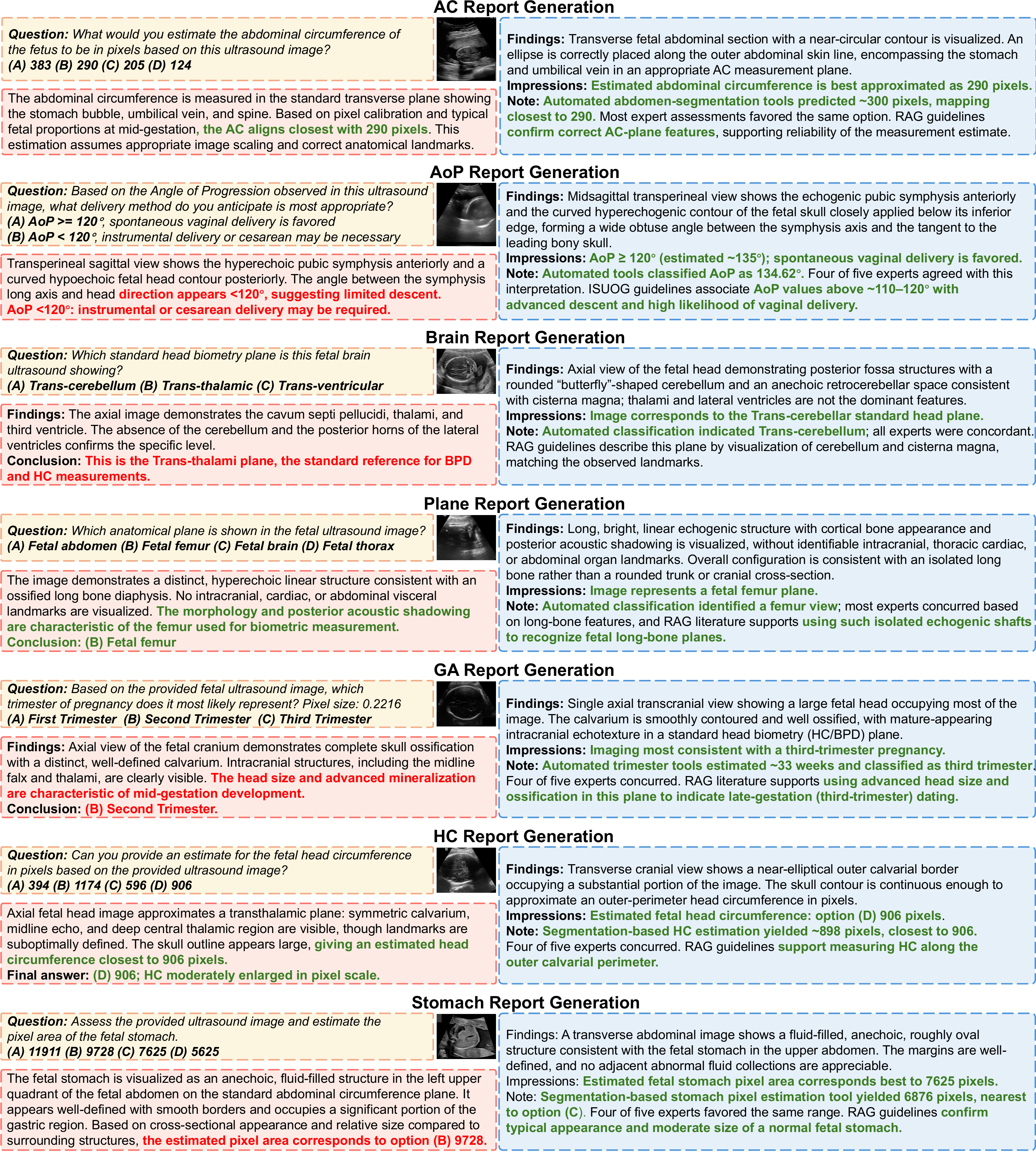}}
\caption{Representative examples of report generation across seven VQA tasks. The yellow boxes show the input VQA questions, the red boxes present the reports generated by the best baseline model, and the blue boxes show the reports generated by FetUSAgents. The undesired and desired responses are highlighted in red and green respectively.}
\label{fig:crop_pic3}
\end{figure}

\begin{table}[t]
\centering
\caption{LLM-based evaluation of report generation quality.
Each score is the average across five dimensions
(Clinical Accuracy, Completeness, Reasoning, Professional Language, and Overall Quality),
normalized relative to GPT-5.4.
For tasks with both binary and multi-class variants,
only the multi-class setting is evaluated.
Best results are in \textbf{bold} and runner-up \underline{underlined}.}
\label{tab:main_results_AI}
\setlength{\tabcolsep}{4pt}
\renewcommand{\arraystretch}{1.12}
\resizebox{\textwidth}{!}{%
\begin{tabular}{@{}lcccccccc@{}}
\toprule

\textbf{Method} & \textbf{AC} & \textbf{AoP} & \textbf{Brain} & \textbf{Plane} & \textbf{GA} & \textbf{HC} & \textbf{Stomach} & \textbf{Avg.} \\
\midrule

\rowcolor{gray!15}
\multicolumn{9}{c}{\textbf{(a) General large language models}} \\
GPT-5.1 & 0.9760 & \underline{1.0290} & 0.8942 & 0.9110 & 1.0101 & 0.8762 & 1.1483 & 0.9778 \\
GPT-5.4 & 1.0000 & 1.0000 & 1.0000 & 1.0000 & 1.0000 & \underline{1.0000} & 1.0000 & 1.0000 \\
Qwen3-Max & \underline{1.0442} & 0.9523 & 0.9870 & 0.6134 & 0.9259 & 0.9187 & 1.2165 & 0.9511 \\
Qwen3-VL-30B-A3B & 1.0303 & 0.9680 & 0.9087 & 0.7384 & 1.0067 & 0.9285 & \underline{1.2502} & 0.9758 \\
DeepSeek-V3.2 & 1.0236 & 0.9404 & 0.7233 & 0.5950 & 0.7499 & 0.9235 & 1.1690 & 0.8750 \\
Doubao-Seed-2.0-Pro & 0.9669 & 0.9212 & 0.5541 & 0.9128 & 0.5009 & 0.9083 & 0.7745 & 0.7912 \\
Gemini-3-Flash-Preview-Thinking & 0.8592 & 0.9852 & \underline{1.1785} & \textbf{1.0709} & \underline{1.0215} & 0.9234 & 1.0834 & \underline{1.0174} \\
\midrule

\rowcolor{gray!15}
\multicolumn{9}{c}{\textbf{(b) Medical large language models}} \\
LLaVA-Med-v1.5-Mistral-7B & 0.8247 & 0.9189 & 0.5038 & 0.5305 & 0.4518 & 0.7334 & 1.0154 & 0.7112 \\
HuatuoGPT-34B & 0.8726 & 0.9323 & 0.8619 & 0.6356 & 0.9211 & 0.8710 & 1.1353 & 0.8900 \\
Hulu-Med-30A3 & 1.0301 & 0.8983 & 0.9026 & 0.8883 & 0.9731 & 0.8441 & 1.1107 & 0.9496 \\
Hulu-Med-32B & 0.9251 & 0.9268 & 0.8780 & 0.8156 & 0.9589 & 0.8705 & 1.1778 & 0.9361 \\
MedGemma-27B & 0.9222 & 0.9479 & 0.8259 & 0.7504 & 0.9553 & 0.9240 & 1.2070 & 0.9332 \\
\midrule

\textbf{FetUSAgents} & \textbf{1.2418} & \textbf{1.1765} & \textbf{1.3138} & \underline{1.0120} & \textbf{1.0743} & \textbf{1.2398} & \textbf{1.2928} & \textbf{1.1930} \\
\bottomrule
\end{tabular}%
}
\end{table}

\begin{table}[t]
\centering
\caption{Expert evaluation of report generation quality.
Each score is the average across four dimensions
(Accuracy, Completeness, Professional Language, and Structural Organization),
normalized relative to GPT-5.4.
For tasks with both binary and multi-class variants,
only the multi-class setting is evaluated.
Best results are in \textbf{bold} and runner-up \underline{underlined}.}
\label{tab:main_results_expert}
\setlength{\tabcolsep}{4pt}
\renewcommand{\arraystretch}{1.12}
\resizebox{\textwidth}{!}{%
\begin{tabular}{@{}lcccccccc@{}}
\toprule

\textbf{Method} & \textbf{AC} & \textbf{AoP} & \textbf{Brain} & \textbf{Plane} & \textbf{GA} & \textbf{HC} & \textbf{Stomach} & \textbf{Avg.} \\
\midrule

\rowcolor{gray!15}
\multicolumn{9}{c}{\textbf{(a) General large language models}} \\
GPT-5.1 & 1.1660 & 1.1156 & 1.0865 & 1.2198 & 1.1045 & 1.0882 & 0.9823 & 1.1090 \\
GPT-5.4 & 1.0000 & 1.0000 & 1.0000 & 1.0000 & 1.0000 & 1.0000 & 1.0000 & 1.0000 \\
Qwen3-Max & 1.1826 & 0.9490 & 1.0163 & 1.1781 & 1.0083 & 1.2243 & 1.1229 & 1.0974 \\
Qwen3-VL-30B-A3B & 1.1676 & 1.0434 & 1.1017 & 1.2611 & 1.0278 & 1.2243 & 1.1250 & 1.1359 \\
DeepSeek-V3.2 & 1.2052 & \underline{1.1701} & 1.0823 & 1.1510 & 1.1205 & 1.1993 & 1.1031 & 1.1474 \\
Doubao-Seed-2.0-Pro & 1.0774 & 1.0104 & 1.1458 & 1.2344 & 1.1747 & 1.0378 & 1.1073 & 1.1125 \\
Gemini-3-Flash-Preview-Thinking & 1.0045 & 1.1330 & \underline{1.2240} & \underline{1.3132} & \underline{1.2076} & 1.1722 & 1.0750 & 1.1614 \\
\midrule

\rowcolor{gray!15}
\multicolumn{9}{c}{\textbf{(b) Medical large language models}} \\
LLaVA-Med-v1.5-Mistral-7B & 0.8913 & 0.9361 & 0.7712 & 0.8229 & 0.5365 & 0.9892 & 0.9715 & 0.8455 \\
HuatuoGPT-34B & 0.9799 & 0.8781 & 0.9410 & 0.9705 & 0.8497 & 0.9014 & 0.8417 & 0.9089 \\
Hulu-Med-30A3 & 1.0785 & 0.8691 & 1.0976 & 1.1585 & 1.1517 & 1.1646 & 0.9427 & 1.0661 \\
Hulu-Med-32B & 1.1378 & 1.0278 & 1.0632 & 1.0488 & 1.0142 & 1.0000 & 0.8333 & 1.0179 \\
MedGemma-27B & \underline{1.2368} & 1.1097 & 1.1587 & 1.2281 & 1.1243 & \textbf{1.2722} & \underline{1.2042} & \underline{1.1906} \\
\midrule

\textbf{FetUSAgents} & \textbf{1.3170} & \textbf{1.3354} & \textbf{1.3306} & \textbf{1.4436} & \textbf{1.2944} & \underline{1.2507} & \textbf{1.2573} & \textbf{1.3184} \\
\bottomrule
\end{tabular}%
}
\end{table}

\begin{figure}[!t]
\centerline{\includegraphics[width=\textwidth]{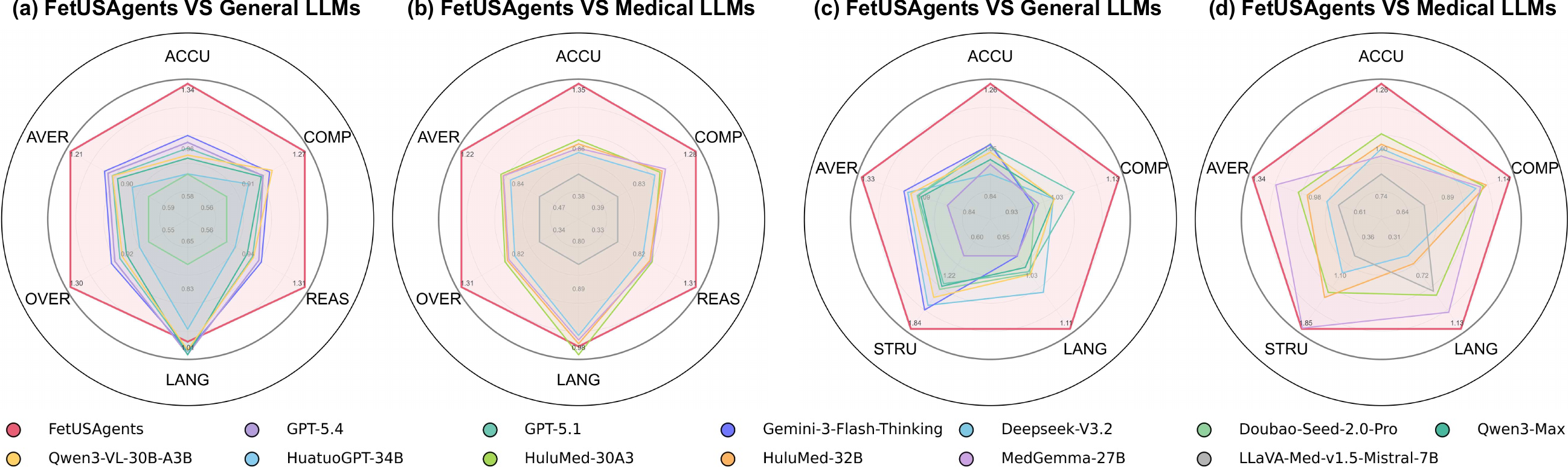}}
\caption{Radar-chart comparison of report generation quality.
(a,\,b) LLM-based and (c,\,d) expert evaluation against
general (a,\,c) and  (b,\,d) LLMs.
ACCU: Clinical Accuracy; COMP: Completeness; REAS: Reasoning;
LANG: Professional Language; OVER: Overall Quality;
STRU: Structural Organization; AVER: Dimension-averaged score.}
\label{fig:crop_pic12}
\end{figure}

\subsection{Comparison with MLLMs on Report Generation}

\begin{figure}[!t]
\centerline{\includegraphics[width=\columnwidth]{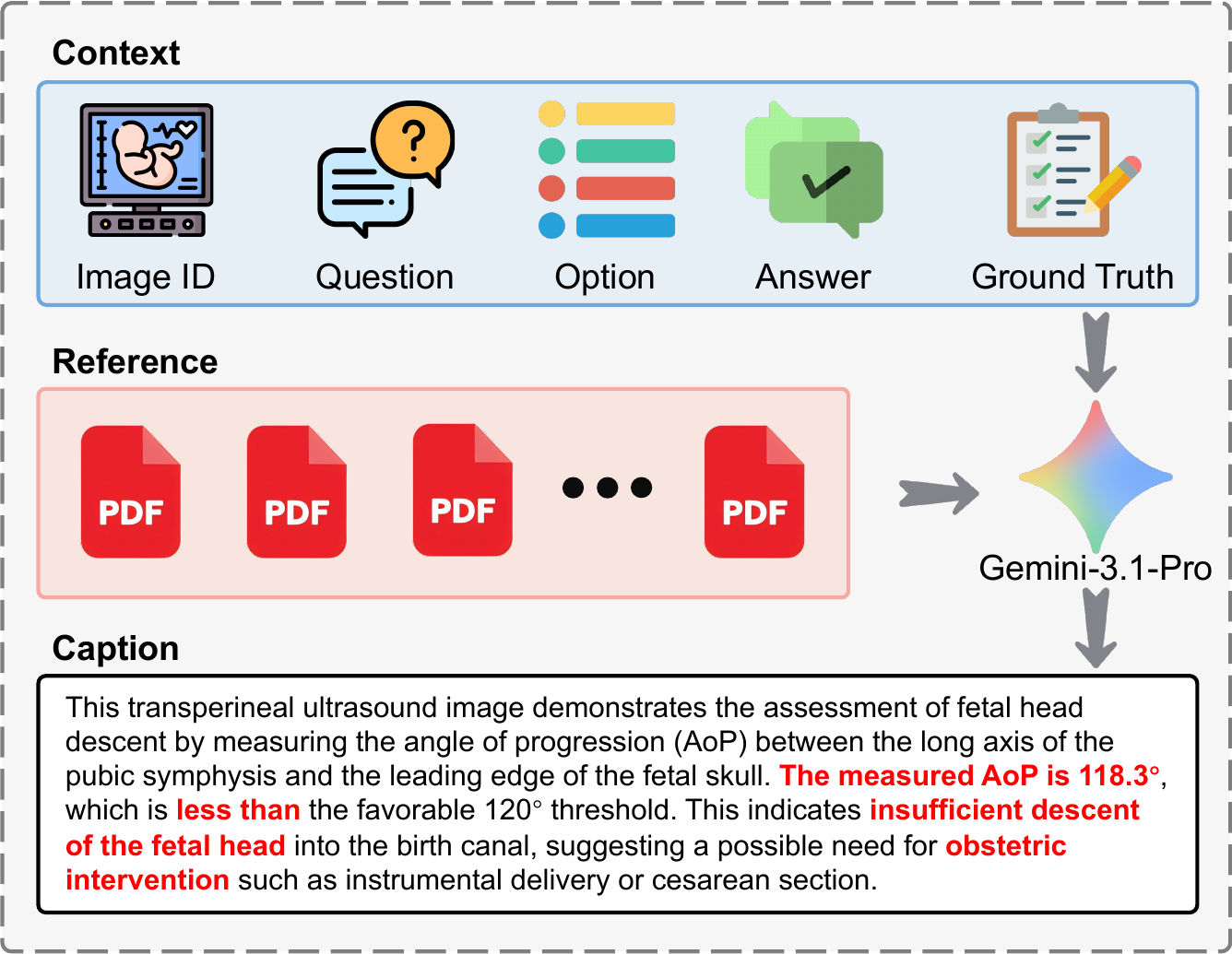}}
\caption{Reference caption construction for LLM-based report evaluation. Task-specific context (image ID, question, options, answer, and ground truth) together with external reference documents are provided to Gemini-3.1-Pro to synthesize clinically grounded reference descriptions.}
\label{fig:crop_pic2}
\end{figure}

Beyond VQA accuracy, report generation quality (Fig.~\ref{fig:crop_pic3})
was assessed from two complementary perspectives:
(i)~LLM-based evaluation, in which Gemini-3-Pro-Thinking\citep{gemini3pro}
scores each report on a 1--5 scale across five dimensions
(i.e., \textit{Clinical Accuracy}, \textit{Completeness}, \textit{Reasoning}, \textit{Professional Language},
and \textit{Overall Quality}), with a reference image description
automatically constructed from contextual information
(Fig.~\ref{fig:crop_pic2}) as auxiliary evaluation context; and
(ii)~expert evaluation, in which two clinical sonographers
independently rate the same reports on a 1--5 scale across four
dimensions (i.e., \textit{Accuracy}, \textit{Completeness}, \textit{Professional Language}, and
\textit{Structural Organization}).
For both evaluations, each model's score is normalized by that of
GPT-5.4 to yield a relative score.

For LLM-Based evaluation, FetUSAgents consistently outperformed all baselines, achieving an
average relative score of 1.1930 (+19.30\% over GPT-5.4), compared
with the strongest general baseline
Gemini-3-Flash-Preview-Thinking (1.0174) and the strongest
medical baseline Hulu-Med-30A3 (0.9496), yielding further
improvements of 17.26\% and 25.63\%, respectively.
Radar chart analysis (Fig.~\ref{fig:crop_pic12}(a,\,b)) showed that
the contour of FetUSAgents consistently enclosed those of all
baselines, with the largest margins on dimensions tied to clinical
grounding and reasoning and a relatively narrow gap on
\textit{Professional Language}, suggesting that current MLLMs can
produce professionally phrased text but remain limited in
integrating domain-specific evidence and performing structured
clinical reasoning.

Expert evaluation further confirmed the superiority of FetUSAgents,
which achieved an average relative score of 1.3184 (+31.84\% over
GPT-5.4), exceeding Gemini-3-Flash-Preview-Thinking (1.1614) and
MedGemma-27B (1.1906) by 13.52\% and 10.73\%, respectively. At the task level, FetUSAgents ranked first on six of seven tasks
and second on HC, demonstrating consistent robustness.
Radar chart analysis (Fig.~\ref{fig:crop_pic12}(c,\,d)) showed that
FetUSAgents again achieved the most balanced profile, with its
contour enclosing all baselines across every expert-rated dimension,
further supporting its clinical reliability for fetal ultrasound
report generation.

\begin{figure}[!h]
\centerline{\includegraphics[width=\textwidth]{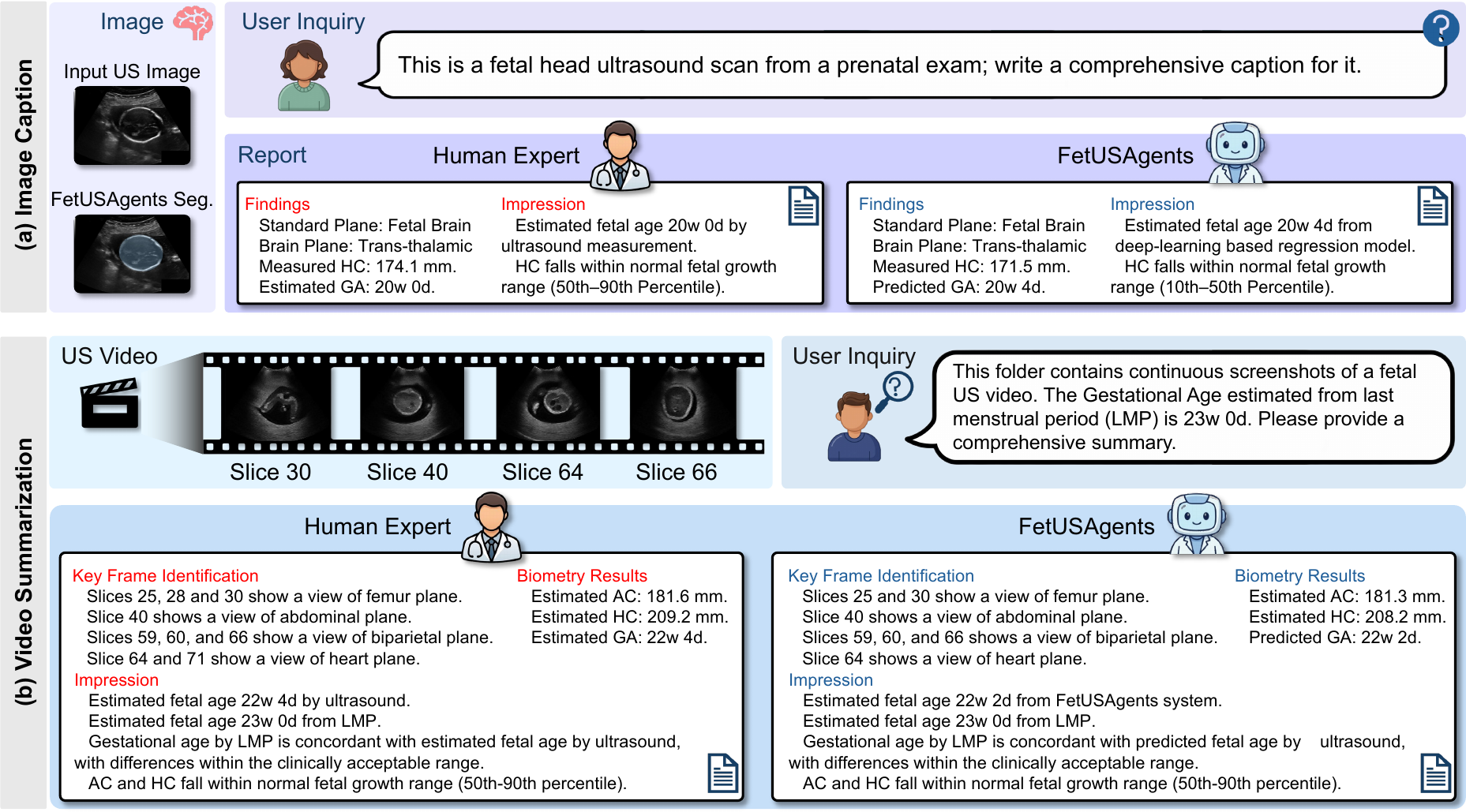}}
\caption{Qualitative examples of general-task outputs. (a)~Image captioning: FetUSAgents identifies the standard plane, performs biometry, and produces a structured report closely aligned with the human expert reference. (b)~Video summarization: FetUSAgents extracts multi-plane keyframes from a continuous scan and cross-validates ultrasound-derived GA against estimates based on the last menstrual period (LMP).}
\label{fig:crop_pic5}
\end{figure}

\subsection{Qualitative Evaluation on General Tasks}
To complement the quantitative analyses above, the end-to-end output of FetUSAgents is qualitatively assessed on two representative general-task scenarios (Fig.~\ref{fig:crop_pic5}). For single-image captioning, FetUSAgents correctly identified the trans-thalamic brain plane, invoked the corresponding Expert Agents to measure HC and estimate GA, and synthesized the anatomical and biometric findings into a structured report closely aligned with the expert reference. For video summarization, FetUSAgents automatically extracted clinically meaningful keyframes from a continuous scan, recognized the corresponding anatomical planes (femur, abdominal, biparietal, and heart views), aggregated frame-wise biometric evidence into a cohesive summary, and cross-validated the ultrasound-derived GA against the LMP-based estimate to verify clinical consistency. Across both scenarios, FetUSAgents exhibited high concordance with expert-provided references in plane identification, quantitative measurements, and final impression formulation, underscoring its capacity to approximate real-world sonographer workflows.

\begin{figure}[!h]
\centerline{\includegraphics[width=\columnwidth]{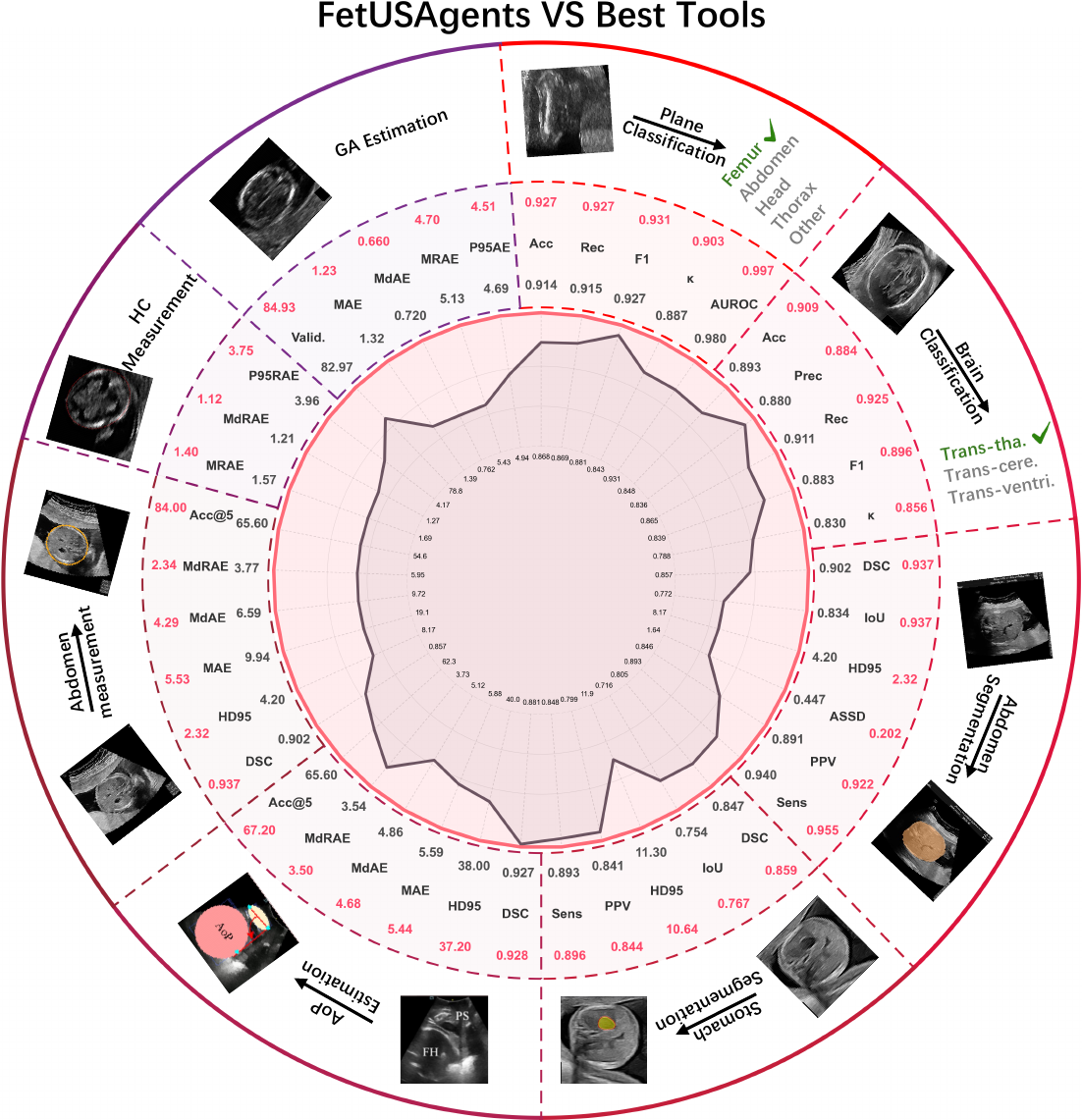}}
\caption{Radar-chart comparison of FetUSAgents (red curves and values) versus the best standalone tool (gray curves and values) across eight fetal ultrasound tasks. All metrics are normalized to a unified scale where the outer boundary consistently indicates superior performance. Abbreviations: Acc, accuracy; Prec, precision; Rec, recall; $\kappa$, Cohen's kappa; DSC, Dice coefficient; IoU, intersection over union; HD95, 95th-percentile Hausdorff distance; ASSD, average symmetric surface distance; PPV, positive predictive value; Sens, sensitivity; MAE/MdAE, mean/median absolute error; MRAE/MdRAE, mean/median relative absolute error; P95AE/P95RAE, 95th-percentile absolute/relative error; Acc@5, accuracy within 5\%; Valid., validity rate.}
\label{fig:crop_pic10}
\end{figure}

\subsection{Comparison with Standalone Tools}

To quantify the gain from multi-tool fusion, each \textit{Expert Agent} was compared with the best-performing standalone tool on the corresponding task across all eight evaluation tasks (Fig.~\ref{fig:crop_pic10}). Task-appropriate metrics were adopted: \textit{Accuracy}, \textit{Precision}, \textit{Recall}, \textit{F1}, \textit{Cohen's}~$\kappa$, and \textit{AUROC} for classification; \textit{DSC}, \textit{IoU}, \textit{HD95}, \textit{ASSD}, \textit{PPV}, and \textit{Sensitivity} for segmentation; \textit{MAE}, \textit{MdAE}, \textit{MRAE/MdRAE}, \textit{P95\,AE/P95\,RAE}, and \textit{Acc@5\%} for biometry, with an additional \textit{Validity Rate} \citep{Fetalclip} for GA estimation. To enable joint visualization of these heterogeneous metrics in a single radar plot, all values were mapped to a unified scale where larger values consistently indicate better performance: for higher-is-better metrics, $x'=x/\max(x_{\mathrm{EA}},\,x_{\mathrm{tool}})$; for lower-is-better metrics, $x'=\min(x_{\mathrm{EA}},\,x_{\mathrm{tool}})/x$, where $x_{\mathrm{EA}}$ and $x_{\mathrm{tool}}$ denote the values of the \textit{Expert Agent} and the best standalone tool, respectively.

Overall, the \textit{Expert Agents} consistently outperformed the best standalone tools on nearly all tasks and metrics. In classification, standard-plane accuracy improved from 0.914 to 0.927 and brain sub-plane $\kappa$ from 0.830 to 0.856, indicating more robust and consistent plane recognition. In segmentation, abdomen DSC rose from 0.902 to 0.937 while HD95 dropped from 4.20 to 2.32, reflecting substantial gains in both regional overlap and boundary precision. In biometry, the largest improvement appeared in AC measurement (MAE: $9.94 \to 5.53$), with GA estimation MAE also decreasing from 1.32 to 1.23. These results confirm that deterministic multi-tool fusion within \textit{Expert Agents} yields more robust and clinically reliable predictions than any individual best model.

\begin{table}[t]
\centering
\caption{Ablation study on VQA accuracy across three representative tasks.}
\label{tab:ablation_acc}
\scriptsize
\setlength{\tabcolsep}{4pt}
\renewcommand{\arraystretch}{1.15}
\begin{tabular}{lcccc}
\toprule
\textbf{Method} & \textbf{Task2} & \textbf{Task6} & \textbf{Task9} & \textbf{Avg} \\
\midrule
w/o Arbiter Agent 
& \underline{0.6719} & 0.8283 & 0.7467 & 0.7490 \\

w/o Expert Agents 
& 0.4063 & \underline{0.8884} & \underline{1.0000} & \underline{0.7649} \\

w/o DPEA  
& 0.5936 & 0.7940 & 0.7467 & 0.7114 \\
\midrule
\textbf{FetUSAgents} 
& \textbf{0.7188} & \textbf{0.9313} & \textbf{1.0000} & \textbf{0.8834} \\
\bottomrule
\end{tabular}
\end{table}

\begin{table}[t]
\centering
\caption{Ablation study on report generation quality across three representative tasks.}
\label{tab:ablation_report}
\scriptsize
\setlength{\tabcolsep}{4pt}
\renewcommand{\arraystretch}{1.15}
\begin{tabular}{lcccc}
\toprule
\textbf{Method} & \textbf{Task2} & \textbf{Task6} & \textbf{Task9} & \textbf{Avg} \\
\midrule
w/o Arbiter Agent 
& \underline{1.1615} & 0.9633 & 1.1420 & \underline{1.0889} \\

w/o Expert Agents 
& 0.9666 & 0.9692 & \textbf{1.2480} & 1.0613 \\

w/o DPEA  
& 1.0113 & 0.9242 & 1.1274 & 1.0210 \\

w/o Evidence bank 
& 1.1529 & \underline{0.9749} & 1.1238 & 1.0839 \\
\midrule
\textbf{FetUSAgents} 
& \textbf{1.1765} & \textbf{1.0120} & \underline{1.2398} & \textbf{1.1428} \\
\bottomrule
\end{tabular}
\end{table}

\subsection{Ablation Study}

To isolate the contribution of each key component, ablation experiments were conducted on three representative tasks (i.e., Task~2, Task~6, and Task~9) spanning measurement, classification, and biometry. Four variants were examined by selectively disabling: (i)~w/o the \textit{Arbiter Agent}, defaulting to majority voting among Voters; (ii)~w/o the \textit{Expert Agents}, using only one tool; (iii)~w/o the DPEA mechanism, collapsing dual-path integration into a single source; and (iv)~w/o the Evidence bank, preventing access to consolidated intermediate evidence.

All four components proved indispensable (Tables~\ref{tab:ablation_acc} and \ref{tab:ablation_report}). Disabling DPEA caused the largest drop (accuracy: $0.8834 \to 0.7114$; report score: $1.1428 \to 1.0210$), confirming that the synergy between deliberative reasoning and computational evidence is the cornerstone of reliable decision-making. Removing \textit{Expert Agents} (accuracy: $0.7649$; report: $1.0613$) and the \textit{Arbiter Agent} (accuracy: $0.7490$; report: $1.0889$) further underscored the necessity of tool-grounded evidence and principled evidence arbitration, respectively. Finally, disabling the Evidence bank lowered the report score to $1.0839$, indicating that aggregated intermediate evidence materially strengthens factual completeness and coherence.

\section{Conclusion}
We introduced FetUSAgents and demonstrated its effectiveness as the first tool-augmented multi-agent system for comprehensive fetal ultrasound interpretation in this study. Our findings reveal that FetUSAgents achieves a VQA accuracy of 0.8791 and an macro-F1 score of 0.8721, surpassing the strongest baseline by over 25\%, indicating its potential to bridge the gap between isolated task-specific predictors and clinically coherent, evidence-driven reasoning. Furthermore, the proposed Dual-Path Evidence Arbitration mechanism and retrieval-enhanced Evidence bank significantly enhance both decision reliability and report generation quality, effectively reducing hallucination while maintaining factual grounding across diverse clinical tasks. Overall, FetUSAgents offers a scalable framework for reliable fetal ultrasound AI, which can serve as a practical clinical assistant for prenatal imaging and inspire future research on agentic systems in highly specialized medical domains.

\bibliographystyle{elsarticle-harv}
\bibliography{refs}{}

\end{document}